\documentclass[sigconf]{acmart}

\pdfoutput=1

\renewcommand\footnotetextcopyrightpermission[1]{} 

\usepackage{booktabs} 

\usepackage{url}
\usepackage[printwatermark]{xwatermark}

\usepackage{microtype}
\newif\iflong
\longfalse

\setcopyright{none}

\acmConference[A Berkeley View of Systems Challenges for AI]{}
\acmPrice{}
\acmDOI{}

\newcounter{ResearchCounter}

\DeclareRobustCommand{\placecounter}[1]{%
\refstepcounter{ResearchCounter}%
\theResearchCounter\label{#1}}

\graphicspath{{figs/}}

\definecolor{codegreen}{rgb}{0,0.6,0}
\definecolor{codegray}{rgb}{0.5,0.5,0.5}
\definecolor{codepurple}{rgb}{0.58,0,0.82}
\definecolor{backcolour}{rgb}{0.95,0.95,0.92}

\begin{document}
\title{A Berkeley View of Systems Challenges for AI}

\author{Ion Stoica, Dawn Song, Raluca Ada Popa, David Patterson, Michael W. Mahoney, Randy Katz, Anthony D. Joseph, Michael Jordan, Joseph M. Hellerstein, Joseph Gonzalez, Ken Goldberg,\\ Ali Ghodsi, David Culler, Pieter Abbeel\footnote{}}\thanks{$^{*}$ Authors listed in reverse alphabetical order.}


\renewcommand{\shortauthors}{}

\begin{abstract}
With the increasing commoditization of computer vision, speech recognition and machine translation systems and the widespread deployment of learning-based back-end technologies such as digital advertising and intelligent infrastructures, AI (Artificial Intelligence) has moved from research labs to production.
%
These changes have been made possible by unprecedented levels of data and computation, by methodological advances in machine learning, by innovations in systems software and architectures, and by the broad accessibility of these technologies.  

The next generation of AI systems promises to accelerate these developments and increasingly impact our lives via frequent interactions and making (often mission-critical) decisions on our behalf, often in highly personalized contexts.
Realizing this promise, however, raises daunting challenges. 
In particular, we need AI systems that make timely and safe decisions in unpredictable environments, that are robust against sophisticated adversaries, and that can process ever increasing amounts of data across organizations and individuals without compromising confidentiality.  These challenges will be exacerbated by the end of the Moore's Law, which will constrain the amount of data these technologies can store and process. In this paper, we propose several open research directions in systems, architectures, and security that can address these challenges and help unlock AI's potential to improve lives and society.
\end{abstract}

%
%



\maketitle

\section{Introduction}
\label{sec:intro}


Conceived in the early 1960's with the vision of emulating human intelligence, AI has evolved towards a broadly applicable engineering discipline in which algorithms and data are brought together to solve a variety of pattern recognition, learning, and decision-making problems. Increasingly, AI intersects with other engineering and scientific fields and cuts across many disciplines in computing. 

In particular, computer \emph{systems} have already proved essential in catalyzing recent progress in AI.  Advances in parallel hardware~\cite{Raina09, Dean12paramserver, TPU} and scalable software systems~\cite{mapreduce, spark, graphlab} have sparked the development of new machine learning frameworks~\cite{theano,Dean12paramserver,mli} and algorithms~\cite{bottou2010sgd, hogwild, adamopt, cocoa} to allow AI to address large-scale, real-world problems. Rapidly decreasing storage costs~\cite{ram-prices,hdd-prices}, crowdsourcing, mobile applications, internet of things (IoT), and the competitive advantage of data~\cite{economistbigdata} have driven further investment in data-processing systems and AI technologies~\cite{forbesaigrowth}. The overall effect is that AI-based solutions are beginning to approach or even surpass human-level capabilities in a range of real-world tasks. 
Maturing AI technologies are not only powering existing industries---including web search, high-speed trading and commerce---but are helping to foster new industries around IoT, augmented reality, biotechnology and autonomous vehicles.

These applications will require AI systems to interact with the real world by making automatic decisions. 
Examples include autonomous drones, robotic surgery, medical diagnosis and treatment, virtual assistants, and many more. As the real world is continually changing, sometimes unexpectedly, these applications need to support \emph{continual or life-long} learning~\cite{thrun1998lifelong,silver2013lifelong} and \emph{never-ending} learning~\citep{mitchell2015nel}. Life-long learning systems aim at solving multiple tasks sequentially by efficiently transferring and utilizing knowledge from already learned tasks to new tasks while minimizing the effect of catastrophic forgetting~\citep{mccloskey1989catastrophic}. Never-ending learning is concerned with mastering a set of tasks in each iteration, where the set keeps growing and the performance on all the tasks in the set keeps improving from iteration to iteration.

Meeting these requirements raises daunting challenges, such as active exploration in dynamic environments, secure and robust decision-making in the presence of adversaries or noisy and unforeseen inputs, the ability to explain decisions, and new modular architectures that simplify building such applications. Furthermore, as Moore's Law is ending, one can no longer count on the rapid increase of computation and storage to solve the problems of next-generation AI systems.

Solving these challenges will require synergistic innovations in  architecture, software, and algorithms. Rather than addressing specific AI algorithms and techniques, this paper examines the essential role that systems will play in addressing  challenges in AI and proposes several promising research directions on that frontier.

\section{What is Behind AI's Recent Success}
\label{sec:behind-ai}


The remarkable progress in AI has been made possible by a ``perfect storm" emerging over the past two decades, bringing together: (1) massive amounts of data, (2) scalable computer and software systems, and (3) the broad accessibility of these technologies. These trends have allowed core AI algorithms and architectures, such as deep learning, reinforcement learning, and Bayesian inference to be explored in problem domains of unprecedented scale and scope.

\subsection{Big data}
\label{sec:bigdata-bigcompute}

With the widespread adoption of online global services, mobile smartphones, and GPS by the end of $1990$s, internet companies such as Google, Amazon, Microsoft, and Yahoo! began to amass huge amounts of data in the form of audio, video, text, and user logs. 
When combined with machine learning algorithms, these massive data sets led to qualitatively better results in a wide range of core services, including classical problems in information retrieval, information extraction, and advertising~\cite{bigdata-effectiveness}. 

\subsection{Big systems}
\label{scalable-systems}

Processing this deluge of data spurred rapid innovations in computer and software systems. To store massive amounts of data, internet service companies began to build massive-scale datacenters, some of which host nearly $100,000$ servers, and provide EB~\cite{youtube-storage} of storage. 
To process this data, companies built new large-scale software systems able to run on clusters of cheap commodity servers. Google developed MapReduce~\cite{mapreduce} and Google File System~\cite{gfs}, followed shortly by the open-source counterpart, Apache Hadoop~\cite{hadoop}. Then came a plethora of systems ~\cite{spark, dryad, graphlab, MuLi, flink},
that aimed to improve  speed, scale, and ease of use. These hardware and software innovations led to the datacenter becoming the new computer~\cite{barroso}.




With the growing demand for machine learning (ML), researchers and practitioners built libraries on top of these systems to satisfy this demand~\cite{mahout,madlib,mllib}. 

The recent successes of deep learning (DL) have spurred a new wave of specialized software systems have emerged to scale out these workloads on CPU clusters and take advantage of specialized hardware, such as GPUs and TPUs. Examples include TensorFlow~\cite{abaditensorflow}, Caffe~\cite{jia2014caffe}, Chainer~\cite{chainer}, PyTorch~\cite{pytorch}, and MXNet~\cite{chen2015mxnet}.

\subsection{Accessibility to state-of-the-art technology}
\label{sec:accessibility}


The vast majority of systems that process data and support AI workloads are built as open-source software, including Spark~\cite{spark}, TensorFlow~\cite{abaditensorflow}, MXNet~\cite{chen2015mxnet}, Caffe~\cite{jia2014caffe}, PyTorch~\cite{pytorch}, and BigDL~\cite{bigdl}. 
Open source allows organizations and individuals alike to leverage state-of-the-art software technology without incurring the prohibitive costs of development from scratch or licensing fees.  

The wide availability of public cloud services (e.g., AWS, Google Cloud, and MS Azure) allows everyone to access virtually unlimited amounts of processing and storage without needing to build large datacenters. Now, researchers can test their algorithms at a moment's notice on numerous GPUs or FPGAs by spending just a few thousands of dollars, which was unthinkable a decade ago.  

\section{Trends and Challenges}
\label{sec:trends-challenges}

%

While AI has already begun to transform many application domains, looking forward, we expect that AI will power a much wider range of services, from health care to transportation, manufacturing to defense, entertainment to energy, and agriculture to retail.
Moreover, while large-scale systems and ML frameworks have already played a pivotal role in the recent success of AI, looking forward, we expect that, together with security and hardware architectures, systems will play an even more important role in enabling the broad adoption of AI.
To realize this promise, however, we need to address significant challenges that are driven by the following trends.

\subsection{Mission-critical AI}
With ongoing advances in AI in applications, from banking to autonomous driving to robot-assisted surgery and to home automation, AI is poised to drive more and more mission-critical applications where human well-being and lives are at stake. 

As AI will increasingly be deployed in dynamic environments, AI systems will need to continually \emph{adapt} and \emph{learn} new ``skills" as the environment changes.
For example, a self-driving car could quickly adapt to unexpected and dangerous road conditions (e.g., an accident or oil on the road), by learning in real time from other cars that have successfully dealt with these conditions. 
Similarly, an AI-powered intrusion-detection system must quickly identify and learn new attack patterns as they happen. In addition, such mission-critical applications must handle noisy inputs and defend against malicious actors.

\emph{\textbf{Challenges:} Design AI systems that learn continually by interacting with a dynamic environment,
while making decisions that are timely, robust, and secure.}

\subsection{Personalized AI} 
From virtual assistants to self-driving cars and  political campaigns, user-specific decisions that take into account user behavior (e.g., a virtual assistant learning a user's accent) and preferences (e.g., a self-driving system learning the level of ``aggressiveness'' a user is comfortable with) are increasingly the focus. While such personalized systems and services provide new functionality and significant economic benefits, they require collecting vast quantities of sensitive personal information and their misuse could affect users' economic and psychological wellbeing.

\emph{\textbf{Challenges:} Design AI systems that enable personalized applications and services yet do not compromise users' privacy and security.} 

\subsection{AI across organizations} 
Companies are increasingly leveraging third-party data to augment their AI-powered services~\cite{cms-sold-data}. Examples include hospitals sharing data to prevent epidemic outbreaks and financial institutions sharing data to improve their fraud-detection capabilities. 
The proliferation of such applications will lead to a transition from data silos---where one company collects data, processes it, and provides the service---to data ecosystems, where applications learns and make decisions using data owned by different organizations.  

\emph{\textbf{Challenges:} Design AI systems that can train on datasets owned by different organizations without compromising their confidentiality, and in the process provide AI capabilities that span the boundaries of potentially competing organization.} 

\subsection{AI demands outpacing the Moore's Law} 
The ability to process and store huge amounts of data has been one of the key enablers of the AI's recent successes (see Section~\ref{sec:bigdata-bigcompute}). However, keeping up with the data being generated will become increasingly difficult due to the following two trends. 

First, data continues to grow exponentially. A 2015 Cisco white paper~\cite{cisco-data-forecast} claims that the amount of data generated by Internet of Everything (IoE) devices by 2018 to be 400ZB, which is almost 50x the estimated traffic in 2015.
According to a recent study~\cite{stephens15}, by $2025$, we will need a three-to-four orders of magnitude improvement in compute throughput to process the aggregate output of all genome sequencers in the world. This would require computation resources to at least double every year. 

Second, this explosion of data is coming at a time when our historically rapidly improving hardware technology is coming to a grinding halt~\cite{hp-architecture-book}. The capacity of DRAMs and disks are expected to double just once in the next decade, and it will take two decades before the performance of CPUs doubles. This slowdown means that storing and processing all generated data will become impracticable.

\emph{\textbf{Challenges:} Develop domain-specific architectures and software systems to address the performance needs of future AI applications in the post-Moore's Law era, including custom chips for AI workloads, edge-cloud systems to efficiently process data at the edge, and techniques for abstracting and sampling data.} 

\section{Research Opportunities}
\label{sec:research-opportunities}

\begin{figure}[h]
\includegraphics[width=80mm]{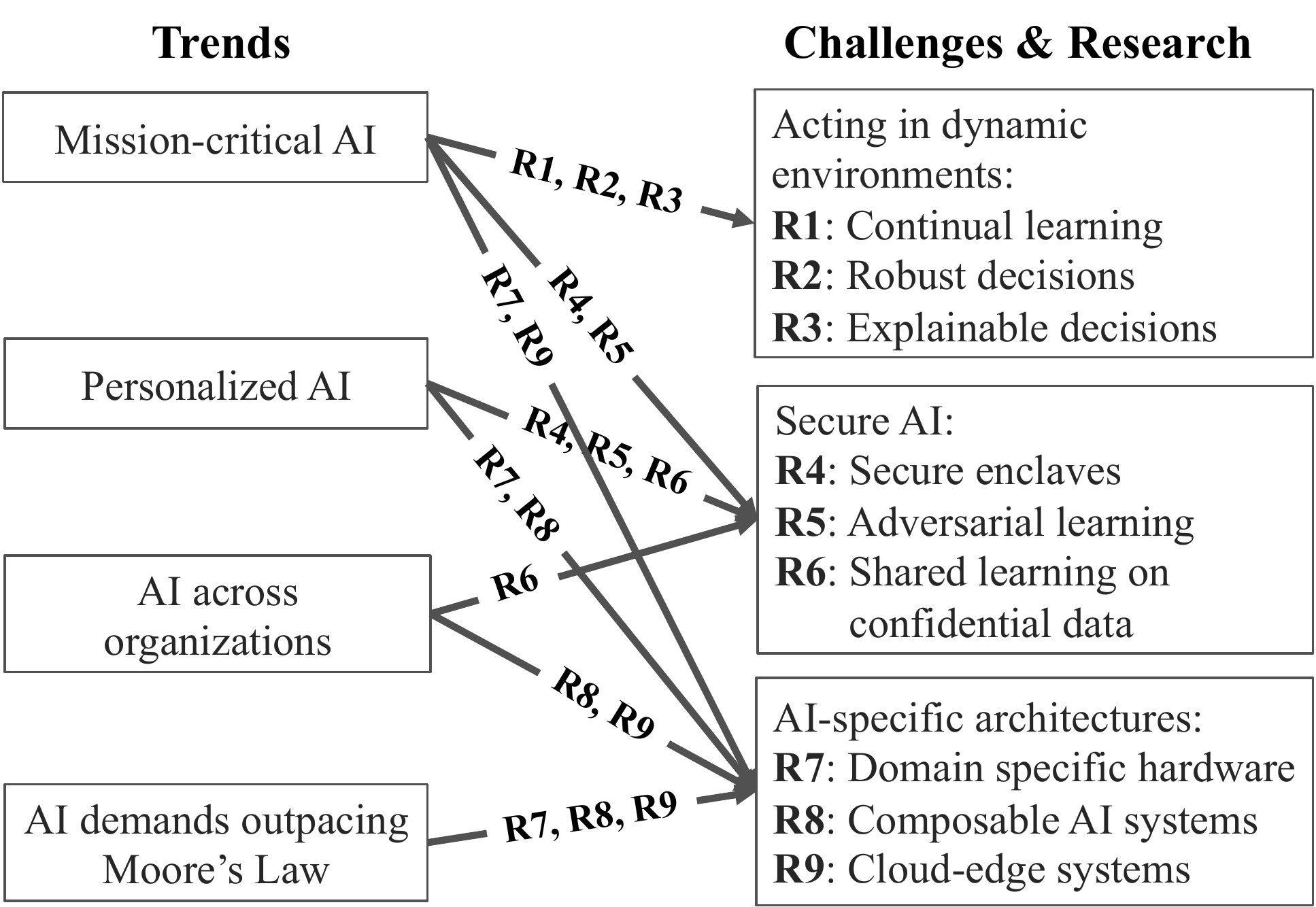}
\vspace{-0.1cm}
\caption{\small{A mapping from trends to challenges and research topics.}}
\label{fig:trends-and-research}
\end{figure}

This section discusses the previous challenges from the systems perspective. In particular, we discuss how innovations in systems, security, and architectures can help address these challenges. We present nine research opportunities (from \textbf{R1} to \textbf{R9}), organized into three topics: acting in dynamic environments, secure AI, and AI-specific architectures. Figure~\ref{fig:trends-and-research} shows the most common relationships between trends, on one hand, and challenges and research topics, on the other hand. 

\vspace{-0.4cm}

 

\subsection{Acting in dynamic environments}
\label{sec:dynamic-env}

Many future AI applications will operate in \emph{dynamic environments}, i.e., 
environments that may change, often rapidly and unexpectedly, and often in non-reproducible ways.
For example, consider a group of robots providing security for an office building.  
When one robot breaks or a new one is added, the other robots must update their strategies for navigation, planning, and control in a coordinated manner. Similarly, when the environment changes, either due to the robots' own actions or to external conditions (e.g., an elevator going out of service, or a malicious intruder), all robots must re-calibrate their actions in light of the change. 
Handling such environments will require AI systems that can react quickly and safely even to scenarios that have not been encountered before. 

\textbf{R\placecounter{ct:cont-learning}: Continual learning.} 
Most of today's AI systems, including movie recommendation, image recognition, and language translation, perform training offline and then make predictions online. 
That is, the learning performed by the system does not happen continually with the generation of the data, but instead it happens sporadicallly, on very different and much slower time scales. 
Typically, models are updated daily, or in the best case hourly, while predictions/decisions happen at second or sub-second granularity. 
This makes them a poor fit for environments that change continually and unexpectedly, especially in mission-critical applications. 
These more challenging environments require agents that \emph{continually learn and adapt} to asynchronous changes.


Some aspects of learning in dynamic environments are addressed by online learning~\cite{online-learning}, in which data arrive temporally and updates to the model can occur as new data arrive. 
However, traditional online learning does not aim to handle control problems,
in which an agent's actions change the environment (e.g., as arise naturally in robotics), nor does it aim to handle cases in which the outcomes of decisions are delayed (e.g., a move in a game of chess whose outcome is only evaluated at the \emph{end}, when the game is lost or won).

These more general situations can be addressed in the framework of Reinforcement Learning (RL). The central task of RL is to learn a function---a ``policy''---that maps observations (e.g., car's camera inputs or user's requested content) to actions (e.g., slowing down the car or presenting an ad) in a sequence that maximizes long-term reward (e.g., avoiding collisions or increasing sales). RL algorithms update the policy by taking into account the impact of agent's actions on the environment, even when delayed. If environmental changes lead to reward changes, RL updates the policy accordingly. RL has a long-standing tradition, with classical success stories including learning to play backgammon at level of the best human players~\cite{tesauro1995temporal}, learning to walk~\cite{tedrake2005learning}, and learning basic motor skills~\cite{peters2008reinforcement}. However, these early efforts require significant tuning for each application. Recent efforts are combining deep neural networks with RL (Deep RL) to develop more robust training algorithms that can work for a variety of environments (e.g., many Atari games~\cite{mnih-dqn-2015}), or even across different application domains, as in the control of (simulated) robots~\cite{schulmanetal_ICML2015} and the learning of robotic manipulation skills~\cite{levine16-e2e}. Noteworthy recent results also include Google's AlphaGo beating the Go world champion~\cite{alphago}, and new applications in medical diagnosis~\cite{rl-medical} and resource management~\cite{rl-dc-cooling}. 

However, despite these successes, 
RL has yet to see widescale real-world application. There are many reasons for this, one of which is that large-scale systems have not been built with these use cases in mind. We believe that the combination of ongoing advances in RL algorithms, when coupled with innovations in systems design, will catalyze rapid progress in RL and drive new RL applications.

\emph{Systems for RL.} 
Many existing RL applications, such as game-playing, rely heavily on simulations, often requiring millions or even billions of simulations to explore the solution space and ``solve" complex tasks.
Examples include playing different variants of a game or experimenting with different control strategies in a robot simulator. 
These simulations can take as little as a few milliseconds, and their durations can be highly variable (e.g., it might take a few moves to lose a game vs.\ hundreds of moves to win one). Finally, real-world deployments of RL systems need to process inputs from a variety of sensors that observe the environment's state, and this must be accomplished under stringent time constraints. Thus, we need systems that can handle arbitrary dynamic task graphs, where tasks are heterogeneous in time, computation, and  resource demands. Given the short duration of the simulations, to fully utilize a large cluster, we need to execute millions of simulations per second. 
None of the existing systems satisfies these requirements. Data parallel systems~\cite{spark,dryad,murray:ciel} handle orders of magnitude fewer tasks per sec, while HPC and distributed DL systems~\cite{abaditensorflow,MXNet,open-mpi} have limited support for heterogeneous and dynamic task graphs. Hence, we need new systems to support effectively RL applications.

\emph{Simulated reality (SR).} 
The ability to interact with the environment is fundamental to RL's success.
Unfortunately, in real-world applications, direct interaction can be slow (e.g., on the order of seconds) and/or hazardous (e.g., risking irreversible physical damage), both of which conflict with the need for having millions of interactions before a reasonable policy is learned. 
%
%
While algorithmic approaches have been proposed to reduce the number of real-world interactions needed to learn policies~\cite{transfer-learning-robotics,transfer-learning-vision, imitation-learning}, 
more generally there is a need for \emph{Simulated Reality (SR)} architectures, in which an agent can continually \emph{simulate and predict} the outcome of the next action before actually taking it
\cite{Sutton}.

SR enables an agent to learn not only much faster but also much more safely. 
Consider a robot cleaning an environment that encounters an object it has not seen before, e.g., a new cellphone. 
The robot could physically experiment with the cellphone to determine how to grasp it, but this may require a long time and might damage the phone. In contrast, the robot could scan the 3D shape of the phone into a simulator, perform a few physical experiments to determine rigidity, texture, and weight distribution, and then use SR to learn how to successfully grasp it without damage. 
%
%

Importantly, SR is quite different from virtual reality (VR); while VR focuses on simulating a hypothetical environment (e.g., Minecraft), sometimes incorporating past snapshots of the real world (e.g., Flight Simulator), SR focuses on continually simulating the physical world with which the agent is interacting. SR is also different from augmented reality (AR), which is primarily concerned with overlaying virtual objects onto real world images.


Arguably the biggest systems challenges associated with SR are to infer continually the simulator parameters in a changing real-world environment and at the same time to run many simulations before taking a single real-time action.  
As the learning algorithm interacts with the world, it gains more knowledge which can be used to improve the simulation. 
Meanwhile, many potential simulations would need to be run between the agent's actions, using both different potential plans and making different ``what-if'' assumptions about the world. 
Thus, the simulation is required to run much faster than real time.

\emph{\textbf{Research:} 
(1) Build systems for RL that fully exploit parallelism, while allowing dynamic task graphs, providing millisecond-level latencies, and running on heterogeneous hardware under stringent deadlines.
(2) Build systems that can faithfully simulate the real-world environment, as the environment changes continually and unexpectedly, and run faster than real time.}

\textbf{R\placecounter{ct:robustness}: Robust decisions.} 
As AI applications are increasingly making decisions on  behalf of humans, notably in mission-critical applications, an important criterion is that they need to be robust to uncertainty and errors in inputs and feedback.
While noise-resilient and robust learning is a core topic in statistics and machine learning, adding system support can significantly improve classical methods.
%
In particular, by building systems that track data provenance, we can diminish uncertainty regarding the mapping of data sources to observations, as well as their impact on states and rewards. We can also track and leverage contextual information that informs the design of source-specific noise models (e.g., occluded cameras).  These capabilities require support for provenance and noise modeling  
in data storage systems.
While some of these challenges apply more generally, two notions of robustness that are particularly important in the context of AI systems and that present particular systems challenges are: (1) robust learning in the presence of noisy and adversarial feedback, and (2) robust decision-making in the presence of unforeseen and adversarial inputs. 

%
Increasingly, learning systems leverage data collected from unreliable sources, possibly with inaccurate labels, and in some cases with deliberately inaccurate labels. 
For example, the Microsoft Tay chatbot relied heavily on human interaction to develop rich natural dialogue capabilities.  
However, when exposed to Twitter messages, Tay quickly took on a dark personality~\cite{tay-wikipedia}. 
%
%

In addition to dealing with noisy feedback, another research challenge is handling inputs for which the system was never trained. 
In particular, one often wishes to detect whether a query input is drawn from a substantially different distribution than the training data, and then take safe actions in those cases. 
An example of a safe action in a self-driving car may be to slow down and stop. 
More generally, if there is a human in the loop, a decision system could 
relinquish control to a human operator. 
Explicitly training models to decline to make predictions for which they are not confident, or to adopt a default safe course of actions, and building systems that chain such models together
can both reduce computational overhead and deliver more accurate and reliable predictions.

\emph{\textbf{Research}: 
\noindent
(1) Build fine grained provenance support into AI systems to connect outcome changes (e.g., reward or state) to the data sources 
that caused these changes, and automatically learn causal, source-specific noise models.  
\noindent
(2) Design API and language support for developing systems that maintain confidence intervals for decision-making, and in particular can flag unforeseen inputs. 
}

\textbf{R\placecounter{ct:explainability}: Explainable decisions.} 
In addition to making black-box predictions and decisions, AI systems will often need to provide explanations for their decisions that are meaningful to humans. 
This is especially important for applications in which there are substantial regulatory requirements  
as well as in applications such as security and healthcare where legal issues arise~\cite{ching17}.  
Here, explainable should be distinguished from interpretable, which is often also of interest.
Typically, the latter means that the output of the AI algorithm is understandable to a subject matter expert in terms of concepts from the domain from which the data are drawn~\cite{CUR_PNAS}, while the former means that one can identify the properties of the input to the AI algorithm that are responsible for the particular output, and can answer counterfactual or ``what-if'' questions.
For example, one may wish to know what features of a particular organ in an X-ray (e.g., size, color, position, form) led to a particular diagnosis and how the diagnosis would change under minor perturbations of those features. Relatedly, one may wish to explore what other mechanisms could have led to the same outcomes, and the relative plausibility of those outcomes. Often this will require not merely providing an explanation for a decision, but also considering other data that could be brought to bear. Here we are in the domain of causal inference, a field which will be essential in many future AI applications, and one which has natural connections to diagnostics and provenance ideas in databases. 

Indeed, one ingredient for supporting explainable decisions is the ability to \emph{record and faithfully replay} the computations that led to a particular decision. 
Such systems hold the potential to help improve decision explainability by replaying a prediction task against past inputs---or randomly or adversarially perturbed versions of past inputs, or more general counterfactual scenarios---to identify what features of the input have caused a particular decision. 
For example, to identify the cause of a false alarm in a video-based security system, one might introduce perturbations in the input video that attenuate the alarm signal (e.g., by masking regions of the image) or search for closely related historical data (e.g., by identifying related inputs) that led to similar decisions.
Such systems could also lead to improved statistical diagnostics and improved training/testing for new models; e.g., by designing models that are (or are not) amenable to explainability.


\emph{\textbf{Research:} 
Build AI systems that can support interactive diagnostic analysis, that faithfully replay past executions, and that can help to determine the features of the input that are responsible for a particular decision, possibly by replaying the decision task against past perturbed inputs. More generally, provide systems support for causal inference.}

\subsection{Secure AI}
\label{sec:secure-ai}

Security is a large topic, many aspects of which will be central to AI applications going forward.
For example, mission-critical AI applications, personalized learning, and learning across multiple organizations all require systems with strong security properties. While there is a wide range of security issues, here we focus on two broad categories.
The first category is an attacker compromising the integrity of the decision process. The attacker can do so either by compromising and taking the control of the AI system itself, or by altering the inputs so that the system will unknowingly render decisions that the attacker wants. The second category is an attacker learning the confidential data on which an AI system was trained on, or learning the secret model. Next, we discuss three promising research topics to defend against such attacks.

\textbf{R\placecounter{ct:enclaves}: Secure enclaves.} 
The rapid rise of public cloud and the increased complexity of the software stack considerably widen the exposure of AI applications to attacks. Two decades ago most applications ran on top of a commercial OS, such as Windows or SunOS, on a single server deployed behind organization's firewalls. Today, organizations run AI applications in the public cloud on a distributed set of servers they do not control, possibly shared with competitors, on a considerably more complex software stack, where the OS itself runs on top of a hypervisor or within a container.
Furthermore, the applications leverage directly or indirectly a plethora of other systems, such as log ingestion, storage, and data processing frameworks. If any of these software components is compromised, the AI applications itself might be compromised.

A general approach to deal with these attacks is providing a ``secure enclave'' abstraction---a secure execution environment---which protects the application running within the enclave from malicious code running outside the enclave. One recent example is Intel's Software Guard Extensions (SGX)~\cite{sgx}, which provides a hardware-enforced isolated execution environment. Code inside SGX can compute on data, while even a compromised operating system or hypervisor (running outside the enclave) cannot see this code or data. SGX also provides remote attestation~\cite{remoteattest}, a protocol enabling a remote client to verify that the enclave is running the expected code. ARM's TrustZone is another example of a hardware enclave. At the other end of the spectrum, cloud providers are starting to offer special bare-bone instances that are physically protected, e.g., they are deployed in secure ``vaults'' to which only authorized personnel, authenticated via fingerprint or iris scanning, has access. 

In general, with any enclave technology, the application developer must trust all the software running within the enclave. Indeed, even in the case of hardware enclaves, if the code running inside the enclave is compromised, it can leak decrypted data or compromise decisions. Since a small code base is typically easier to secure, one research challenge is to split the AI system's code into code running inside the enclave, hopefully as little as possible, and code running outside of the enclave, in untrusted mode, by leveraging cryptographic techniques. Another approach to ensure that code inside the enclave does not leak  sensitive information is  to develop static and dynamic verification tools as well as sandboxing~\cite{baumann2015shielding,schuster2015vc3,arnautov2016scone}.

Note that beside minimizing the trusted computing base, there are two additional reasons for splitting the application code: increased functionality and reduced cost. First, some of the functionality might not be available within the enclave, e.g., GPU processing for running Deep Learning (DL) algorithms, or services and applications which are not vetted/ported yet to run within the secure enclave. 
Second, the secure instances offered by a cloud provider can be significantly more expensive than regular instances.

\emph{\textbf{Research:} Build AI systems that leverage secure enclaves to ensure data confidentiality, user privacy and decision integrity, possibly by splitting the AI system's code between a minimal code base running within the enclave, and code running outside the enclave. Ensure the code inside the enclave does not leak information, or compromise decision integrity.}

\textbf{R\placecounter{ct:adversarial-learning}: Adversarial learning.} 
The adaptive nature of ML algorithms opens the learning systems to new categories of attacks that aim to compromise the integrity of their decisions by maliciously altering training data or decision input. There are two broad types of attacks: \emph{evasion attacks} and \emph{data poisoning attacks}. 

Evasion attacks happen at the inference stage,  where an adversary attempts to craft data that is incorrectly classified by the learning system~\cite{szegedy2013intriguing,goodfellow2014explaining}. An example is altering the image of a stop sign slightly such that it still appears to a human to be a stop sign but is  seen by an autonomous vehicle as a yield sign. 

Data poisoning attacks happen at the training stage, where an adversary injects poisoned data (e.g., data with wrong labels) into the training data set that cause the learning system to learn the wrong model, such that the adversary thereby has input data incorrectly classified by the learner~\cite{mei2015security,xiao2015feature,mei2015using}. Learning systems that are periodically retrained to handle non-stationary input data are particularly vulnerable to this attack, if the weakly labeled data being used for retraining is collected from unreliable or untrustworthy sources. With new AI systems continually learning by interacting with dynamic environments, handling data poisoning attacks becomes increasingly important. 

Today, there are no effective solutions to protect against evasion attacks. As such, there are a number of open research challenges: provide better understanding of why adversarial examples are often easy to find, investigate what method or combination of different methods may be effective at defending against adversarial examples, and design and develop systematic methods to evaluate potential defenses. For data poisoning attacks, open challenges include how to detect poisoned input data and how to build learning systems that are resilient to different types of data poisoning attacks. 
In addition, as data sources are identified to be fraudulent or explicitly retracted for regulatory reasons, we can leverage replay (see R\ref{ct:explainability}: Explainable decisions) and incremental computation to efficiently eliminate the impact of those sources on learned models.  As pointed out previously, this ability is enabled by combining modeling with provenance and efficient computation in data storage systems.

\emph{\textbf{Research:} Build AI systems that are robust against adversarial inputs both during training and prediction (e.g., decision making), possibly by designing new machine learning models and network architectures, leveraging provenance to track down fraudulent data sources, and replaying to redo decisions after eliminating the fraudulent sources.}

\textbf{R\placecounter{ct:shared-learning}: Shared learning on confidential data.} 
Today, each company typically collects data individually, analyzes it, and uses this data to implement new features and products. However, not all organizations possess the same wealth of data as found in the few large AI-focused corporations, such as Google, Facebook, Microsoft, and Amazon. Going forward, we expect more and more organizations to collect valuable data,  more third-party data services to be available, and more benefit to be gained from learning over data from multiple organizations (see Section~\ref{sec:trends-challenges}). 

Indeed, from our own interaction with industry, we are learning about an increasing number of such scenarios. 
A large bank provided us with a scenario in which they and other banks would like to pool together their data and use shared learning to improve their collective fraud detection algorithms. While these banks are natural competitors in providing financial services, such "cooperation" is critical to minimize their losses due to fraudulent activities. A very large health provider described a similar scenario in which competing hospitals would like to share data to train a shared model predicting flu outbreaks without sharing the data for other purposes. This would allow them to improve the response to epidemics and contain the outbreaks, e.g., by rapidly deploying mobile vaccination vans at critical locations.  At the same time,  every hospital must protect the confidentiality of their own patients.

The key challenge of shared learning is how to learn a model on data belonging to different (possible competitive) organizations without leaking relevant information about this data during the training process. One possible solution would be to pool all the data in a hardware enclave and then learn the model. However, this solution is not always feasible as hardware enclaves are not yet deployed widely, and, in some cases, the data cannot be moved due to regulatory constraints or its large volume. 

Another promising approach is using \emph{secure multi-party computation (MPC)}~\cite{goldreich1987play,ben1988completeness,malkhi2004fairplay}. MPC enables $n$ parties, each having a private input, to compute a joint function over the input without any party learning the inputs of the other parties. Unfortunately, while MPC is effective for simple computations, it has a nontrivial overhead for complex computations, such as model training. An interesting research direction is how to partition model training into (1) local computation and (2) computation using MPC, so that we minimize the complexity of the MPC computation. 

While training a model without compromising data confidentiality is a big step towards enabling shared learning, unfortunately, it is not always enough. Model serving---the inferences (decisions) rendered based on the model---can still leak information about the data~\cite{shokri2016membership,fredrikson2015model}. 
One approach to address this challenge is \emph{differential privacy}~\cite{DBLP:conf/icalp/Dwork06,dwork2008differential,dwork2014algorithmic}, a popular technique  proposed in the context of statistical databases. Differential privacy adds noise to each query to protect the data privacy, hence effectively trading accuracy for privacy~\cite{duchi-etal}. A central concept of differential privacy is the privacy budget which caps the number of queries given a privacy guarantee. 

There are three interesting research directions when applying differential privacy to model serving. 
First, a promising approach is to leverage differential privacy for complex models and inferences, by taking advantage of the inherent statistical nature of the models and predictions. Second, despite the large volume of theoretical research, there are few practical differential privacy systems in use today. An important research direction is to build tools and systems to make it easy to enable differential privacy for real-world applications, including intelligently selecting which privacy mechanisms to use for a given application and automatically converting non-differentially-private computations to differentially-private computations. Finally, one particular aspect in the context of continual learning is that data privacy can be time dependent, that is, the privacy of fresh data is far more important than the privacy of historical data. 
Examples are stock market and online bidding, where the privacy of the fresh data is paramount, while the historical data is sometimes publicly released. This aspect could enable the development of new differential privacy systems with adaptive privacy budgets that apply only to decisions on the most recent data. Another research direction is to further develop the notion of differential privacy under continuous observation and data release~\cite{DBLP:conf/icalp/ChanSS10,dwork2010differential}. 

Even if we are able to protect data confidentiality during training and decision making, this might still not be enough. Indeed, even if confidentiality is guaranteed, an organization might refuse to share its data for improving a model from which its competitors might benefit. Thus, we need to go beyond guaranteeing confidentiality and provide \emph{incentives} to organizations to share their data or byproducts of their data. 
Specifically, we need to develop approaches that ensure that by sharing data, an organization gets strictly better service (i.e., better decisions) than not sharing data. This requires ascertaining the quality of the data providing by a given organization---a problem which can be tackled via a ``leave-one-out'' approach in which performance is compared both with and without that organization's data included in the training set. We then provide decisions that are corrupted by noise at a level that is inversely proportional to the quality of the data provided by an organization. This incentivizes an organization to provide higher-quality data.  Overall, such incentives will need to be placed within a framework of mechanism design to allow organizations to forge their individual data-sharing strategies.

\emph{\textbf{Research:} Build AI systems that (1) can learn across multiple data sources without leaking information from a data source during training or serving, and (2) provide incentives to potentially competing organizations to share their data or models.}

\subsection{AI-specific architectures}
\label{sec:ai-architectures}

AI demands will drive innovations both in systems and hardware architectures. These new architectures will aim not only to improve the performance, but to simplify the development of the next generation of AI applications by providing rich libraries of modules that are easily composable. 

\textbf{R\placecounter{ct:domain-specific}: Domain specific hardware.} 
The ability to process and store huge amounts of data has been one of the key enablers of the AI's recent successes (see Section~\ref{sec:bigdata-bigcompute}). 
However, continuing to keep up with the data being generated will be increasingly challenging. As discussed in Section~\ref{sec:trends-challenges}, 
while data continues to grow exponentially, the corresponding performance-cost-energy improvements that have fueled the computer industry for more than $40$ years are reaching the end-of-line: 
\begin{itemize}
\item Transistors are not getting much smaller due to the ending of Moore's Law,
\item Power is limiting what can be put on a chip due to the end of Dennard scaling,
\item We've already switched from one inefficient processor/chip to about a dozen efficient processors per chip, but there are limits to parallelism due to Amdahl's Law.
\end{itemize}

The one path left to continue the improvements in performance-energy-cost of processors is developing domain-specific processors. These processors do only a few tasks, but they do them extremely well. Thus, the rapid improvements in processing that we have expected in the Moore's law era must now come through innovations in computer architecture instead of semiconductor process improvements. Future servers will have much more heterogeneous processors than in the past. One trailblazing example that spotlights domain specific processors is Google's Tensor Processing Unit, which has been deployed in its datacenters since 2015 and is regularly used by billions of people. It performs the inference phase of deep neural networks $15\times$ to $30\times$ faster than its contemporary CPUs and GPUs and its performance per watt is $30\times$ to $80\times$ better. In addition, Microsoft has announced the availability of FPGA-powered instances on its Azure cloud~\cite{catapult}, and a host of companies, ranging from Intel to IBM, and to startups like Cerebras and Graphcore are developing specialized hardware for AI that promise orders of magnitude performance improvements over today's state-of-the-art processors~\cite{cerebras,graphcore,nervana,truenorth}.


With DRAM subject to the same limitations, there are several novel technologies being developed that hope to be its successor. 3D XPoint from Intel and Micron aims to provide $10\times$ storage capacity with DRAM-like performance. STT MRAM aims to succeed Flash, which may hit similar scaling limits as DRAM. Hence, the memory and storage of the cloud will likely have more levels in the hierarchy and contain a wider variety of technologies. Given the increasing diversity of processors, memories, and storage devices, mapping services to hardware resources will become an even more challenging problem. These dramatic changes suggest building cloud computing from a much more flexible building block than the classic standard rack containing a top-of-rack switch and tens of servers, each with $2$ CPU chips, $1$ TB of DRAM, and $4$ TBs of flash. 

For example, the UC Berkeley Firebox project~\cite{firebox} proposes a multi-rack supercomputer that connects thousands of processor chips with thousands of DRAM chips and nonvolatile storage chips using fiber optics to provide low-latency, high-bandwidth, and long physical distance. 
Such a hardware system would allow system software to provision computation services with the right ratio and type of domain-specific processors, DRAM, and NVRAM. Such resource disaggregation at scale would significantly improve the allocation of increasingly diverse tasks to correspondingly heterogeneous resources. It is particularly valuable for AI workloads, which are known to gain significant performance benefits from large memory and have diverse resource requirements that don't all conform to a common pattern.

Besides performance improvements, new hardware architectures will also provide additional functionality, such as security support.  While Intel's SGX and ARM's TrustZone are paving the way towards hardware enclaves, much more needs to be done before they can be fully embraced by AI applications. In particular, existing enclaves exhibit various resource limitations such as addressable memory, and they are only available for a few general purpose CPUs. Removing these limitations, and providing a uniform hardware enclave abstraction across a diverse set of specialized processors, including GPUs and TPUs, are promising directions of research. In addition, open instruction set processors, such as RISC-V represent an exciting ``playground'' to develop new security features.

\emph{\textbf{Research:} (1) Design domain-specific hardware architectures to improve the performance and reduce power consumption of AI applications by orders of magnitude, or enhance the security of these applications. (2) Design AI software systems to take advantage of these domain-specific architectures, resource disaggregation architectures, and future non-volatile storage technologies.}


\textbf{R\placecounter{ct:composable-ai}: Composable AI systems.} 
\label{sec:composable-ai}
Modularity and composition have played a fundamental role in the rapid progress of software systems, as they allowed developers to rapidly build and evolve new systems from existing components. Examples range from microkernel OSes~\cite{Accetta86mach,liedtke95}, LAMP stack~\cite{lamp-stack}, microservice architectures~\cite{microservices}, and the internet~\cite{Clark88}. In contrast, today's AI systems are monolithic which makes them hard to develop, test, and evolve.

Similarly, modularity and composition will be key to increasing development speed and adoption of AI, by making it easier to integrate AI in complex systems. Next, we discuss several research problems in the context of model and action composition. 

\emph{Model composition} 
is critical to the development of more flexible and powerful AI systems. Composing multiple models and querying them in different patterns enables a tradeoff between decision accuracy, latency, and throughput in a model serving system
~\cite{clipper,tensorflow-serving}
In one example, we can query models serially, where each model either renders the decision with sufficiently high accuracy or says ``I'm not sure''.  
In the latter case, the decision is passed to the next model in the series. By ordering the models from the highest to the lowest ``I'm not sure'' rate, and from lowest to the highest latency, we can optimize both latency and accuracy. 


To fully enable model composition, many challenges remain to be addressed. Examples are (1) designing a declarative language to capture the topology of these components and specifying performance targets of the applications, (2) providing accurate performance models for each component, including resource demands, latency and throughput, and (3) scheduling and optimization algorithms to compute the execution plan across components, and map components to the available resources to satisfy latency and throughput requirements while minimizing costs. 


\emph{Action composition} consists of aggregating sequences of basic decisions/actions into coarse-grained primitives, also called \emph{options}. In the case of a self-driving car, an example of an option is changing the lane while driving on a highway, while the actions are speeding up, slowing down, turning left or right, signaling the change of direction, etc. In the case of a robot, an example of a primitive could be grasping an object, while actions include actuating the robot's joints. Options have been extensively studied in the context of hierarchical learning~\cite{dayanH92, dietterich98, parrR97, singh92a, suttonPS99, thrunS94}. Options can dramatically speed up learning or adaptation to a new scenario by allowing the agent to select from a list of existing options to accomplish a given task, rather than from a much longer list of low-level actions. 

A rich library of options would enable the development of new AI applications by simply composing the appropriate options the same way web programmers develop applications today in just a few lines of code by invoking powerful web APIs. In addition, options can improve responsiveness as selecting the next action within an option is a much simpler task than selecting an action in the original action space. 

\emph{\textbf{Research:} Design AI systems and APIs that allow the composition of models and actions in a modular and flexible manner, and develop rich libraries of models and options using these APIs to dramatically simplify the development of AI applications. }

\textbf{R\placecounter{ct:cloud-edge}: Cloud-edge systems.} 
Today, many AI applications such as speech recognition and language translation are deployed in the cloud. Going forward we expect a rapid increase in AI systems that span edge devices and the cloud. 
On one hand, AI systems which are currently cloud only, such as user recommendation systems~\cite{federatedlearning}, are moving some of their functionality to edge devices to improve security, privacy, latency and safety (including the ability to cope with being disconnected from the internet).
On the other hand, AI systems  currently deployed at the edge, such as self-driving cars, drones, and home robots, are increasingly sharing data and leveraging the computational resources available in the cloud to update models and policies~\cite{cloud-robotics-survey}.

However, developing cloud and the cloud-edge systems is challenging for several reasons. First, there is a large discrepancy between the capabilities of edge devices and datacenter servers. We expect this discrepancy to increase in the future, as edge devices, such as cellphones and tablets, have much more stringent power and size constraints than servers in datacenters. Second, edge devices are extremely heterogeneous both in terms of resource capabilities, ranging from very low power ARM or RISC-V CPUs that power IoT devices to powerful GPUs in self-driving cars, and software platforms. This heterogeneity makes application development much harder. Third, the hardware and software update cycles of edge devices are significantly slower than in a datacenter. Fourth, as the increase in the storage capacity slows down, while the growth in the data being generated continues unabated, it may no longer be feasible or cost effective to store this deluge of data. 


There are two general approaches to addressing the mix of cloud and edge devices. The first is to repurpose code to multiple heterogeneous platforms via retargetable software design and compiler technology. To address the wide heterogeneity of edge devices and the relative difficulty of upgrading the applications running on these devices, we need new software stacks that abstract away the heterogeneity of devices by exposing the hardware capabilities to the application through common APIs. Another promising direction is developing compilers and just-in-time (JIT) technologies to efficiently compile on-the-fly complex algorithms and run them on edge devices. This approach can leverage recent code generation tools, such as TensorFlow's XLA~\cite{tensorflow-xla}, Halide~\cite{halide}, and Weld~\cite{weld}.

The second general approach is to design AI systems that are well suited to partitioned execution across the cloud and the edge. As one example, model composition (see Section~\ref{sec:composable-ai}) could allow one to run the lighter but less accurate models at the edge, and the computation-intensive but higher-accuracy models in the cloud.
This architecture would improve decision latency, without compromising accuracy, and it has been already employed in recent video recognition systems~\cite{noscope,liveva}. In another example, action composition would allow building systems where learning of hierarchical options~\cite{ddco-F17} takes place on powerful clusters in the cloud, and then execution of these options happens at the edge.

Robotics is one application domain that can take advantage of a modular cloud-edge architecture. Today, there is a scarcity of open source platforms to develop robotic applications. ROS, arguably the most popular such platform in use today, is confined to running locally and lacks many performance optimizations required by real-time applications. To take advantage of the new developments in AI research such as shared and continual learning, we need systems that can span both edge devices (e.g., robots) and the cloud. 
Such systems would allow developers to seamlessly migrate the functionality between a robot and the cloud to optimize decision latency and learning convergence. While the cloud can run sophisticated algorithms to continually update the models by leveraging the information gathered by distributed robots in real time, the robots can continue to execute the actions locally based on previously downloaded policies. 
 
To address the challenge of the data deluge collected by the edge devices, learning-friendly compression methods can be used to reduce processing overhead.  Examples of such methods include sampling and sketching, which have already been successfully employed for analytics workloads~\cite{cormode2012synopses,olken1990random,Hellerstein97,blinkdb,macrobase}. One research direction is to aggressively leverage sampling and sketching in a systematic way to support a variety of learning algorithms and prediction scenarios. An arguably more difficult challenge is to reduce the storage overhead, which might require to delete data. The key challenge here is that we do not always know how the data will be used in the future. This is essentially a compression problem, but compression for the purposes of ML algorithms. Again, distributed approaches based in materialized samples and sketches can help provide solutions to this problem, as can ML-based approaches in the form of feature selection or model selection protocols.

\emph{\textbf{Research:} Design cloud-edge AI systems that (1) leverage the edge to reduce latency, improve safety and security, and implement intelligent data retention techniques, and (2) leverage the cloud to share data and models across edge devices, train sophisticated computation-intensive models, and take high quality decisions. }


\section{Conclusion}
\label{sec:conclusion}

The striking progress of AI during just the last decade is leading 
to the successful transition from the research lab into commercial services that have previously required human input and oversight.  Rather than replacing human workers, AI systems and robots have potential to enhance human performance and facilitate new forms of collaboration ~\cite{Multiplicity-June-2017}.

To realize the full promise of AI as a positive force in our lives, 
there are daunting challenges to overcome, and many of these challenges are related to systems and infrastructure. These challenges are driven by the realization that AI systems will need to make decisions that are faster, safer, and more explainable, securing these decisions as well as the learning processes against ever more sophisticated types of attacks, continuously increasing the computation capabilities in the face of the end of Moore's Law, and building composable systems that are easy to integrate in existing applications and can span the cloud and the edge. 


This paper proposes several open research directions in systems, architectures, and security that have potential to address these challenges. We hope these questions will inspire new research that can advance AI and make it more capable, understandable, secure and reliable.

\bibliographystyle{ACM-Reference-Format}
\bibliography{references}

\end{document}